\documentclass[conference]{IEEEtran}
\IEEEoverridecommandlockouts
\usepackage{cite}
\usepackage{amsmath,amssymb,amsfonts}
\usepackage{algorithmic}
\usepackage{graphicx}
\usepackage{textcomp}
\usepackage{xcolor}
\usepackage{booktabs}
\usepackage{multirow} 
\usepackage{caption} 
\def\BibTeX{{\rm B\kern-.05em{\sc i\kern-.025em b}\kern-.08em
    T\kern-.1667em\lower.7ex\hbox{E}\kern-.125emX}}
\begin{document}

\title{Measuring and Controlling the Spectral Bias for Self-Supervised Image Denoising}

\author{\IEEEauthorblockN{1\textsuperscript{st} Wang Zhang\thanks{These authors contributed equally.}, Huaqiu Li$^{*}$
\IEEEauthorblockA{\textit{Tsinghua University}}
}
\and
\IEEEauthorblockN{2\textsuperscript{nd} Xiaowan Hu\IEEEauthorblockA{\textit{Tsinghua University}}}
\and
\IEEEauthorblockN{3\textsuperscript{nd} Tao Jiang\IEEEauthorblockA{\textit{Tsinghua University}}}
\and
\IEEEauthorblockN{4\textsuperscript{nd} Zikang Chen\IEEEauthorblockA{\textit{Tsinghua University}}}
\and
\IEEEauthorblockN{Haoqian Wang\dag \thanks{Haoqian Wang is corresponding author. wanghaoqian@tsinghua.edu.cn}\IEEEauthorblockA{\textit{Tsinghua University}}}

}
    





\maketitle

\begin{abstract}
Current self-supervised denoising methods for paired noisy images typically involve mapping one noisy image through the network to the other noisy image. However, after measuring the spectral bias of such methods using our proposed Image Pair Frequency-Band Similarity, it suffers from two practical limitations. Firstly, the high-frequency structural details in images are not preserved well enough. Secondly, during the process of fitting high frequencies, the network learns high-frequency noise from the mapped noisy images. To address these challenges, we introduce a \textbf{S}pectral \textbf{C}ontrolling network (SCNet) to optimize self-supervised denoising of paired noisy images. First, we propose a selection strategy to choose frequency band components for noisy images, to accelerate the convergence speed of training. Next, we present a parameter optimization method that restricts the learning ability of convolutional kernels to high-frequency noise using the Lipschitz constant, without changing the network structure. Finally, we introduce the \textbf{S}pectral \textbf{S}eparation and low-rank \textbf{R}econstruction module (SSR module), which separates noise and high-frequency details through frequency domain separation and low-rank space reconstruction, to retain the high-frequency structural details of images. Experiments performed on synthetic and real-world datasets
 verify the effectiveness of SCNet. The code will be released soon.
\end{abstract}


\section{Introduction}
\label{sec:intro}

Image denoising is a fundamental task in low-level image processing, aiming to remove noise and restore clean images. Image sensors introduce various types of noise such as thermal, photon, and dark current due to factors like the working environment and electronic components. 
Importance of image denoising lies in visual applications, where the quality of denoising significantly impacts the performance of downstream tasks such as super-resolution
\cite{Huang_Lu_Shao_Ran_Zhou_Fang_Zhang_2019}, semantic segmentation \cite{Liu_Wen_Jiao_Liu_Wang_Huang_2020}, and object detection \cite{B._Tom_George_2019}. 
With the advance of neural networks, 
supervised denoising methods, such as DnCNN \cite{B._Tom_George_2019}, NBNet \cite{Cheng_Wang_Huang_Liu_Fan_Liu_2021}, and NAFNet \cite{chen2022simple}, have proven effective in removing noise. However, these methods rely on large quantities of high-quality noisy-clean image pairs.
\begin{figure}[t]
\centering
\begin{minipage}[b]{0.46\columnwidth}
    \centering
    \includegraphics[width=\linewidth]{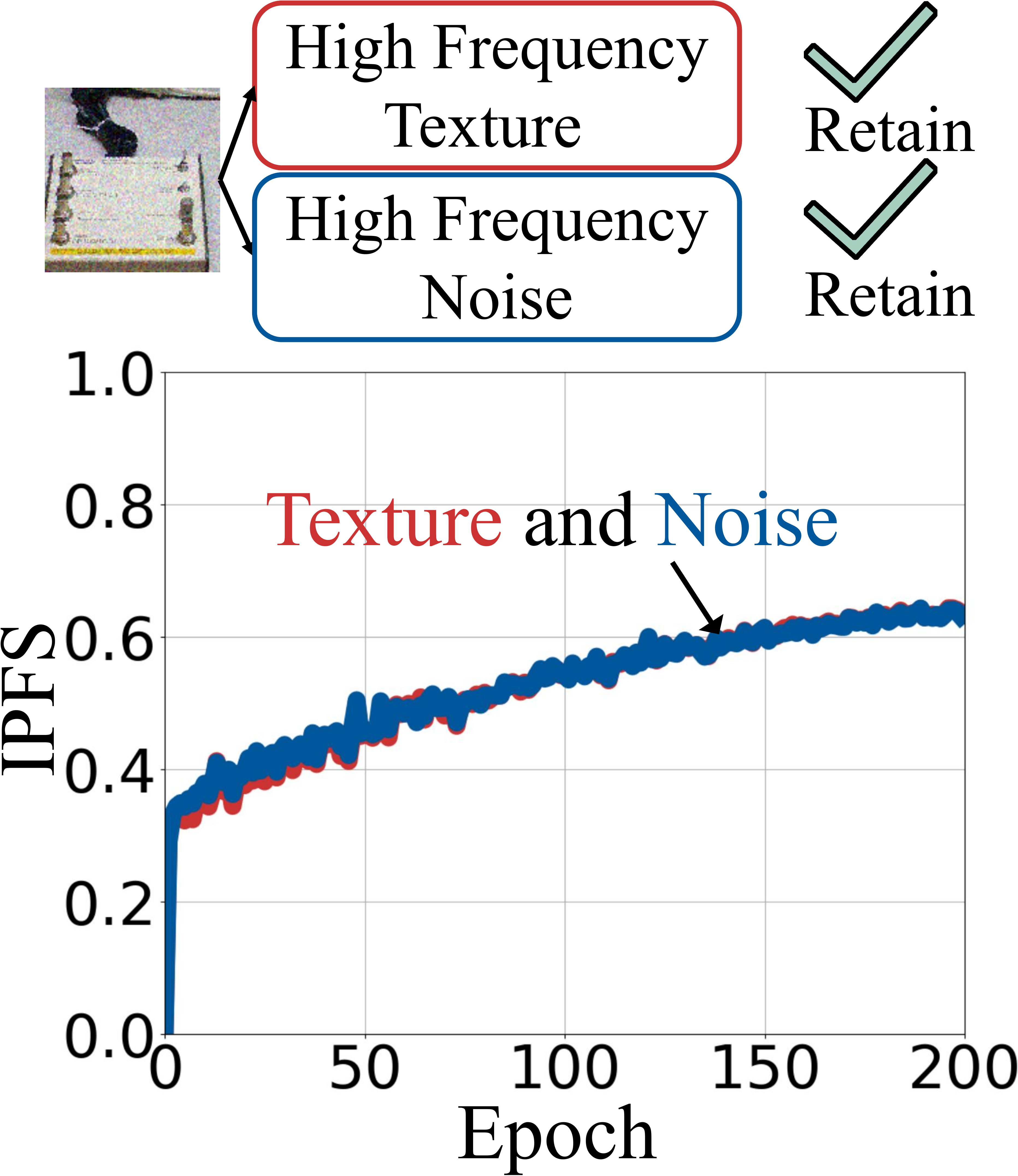} 

    \caption*{(a) Existing method} 
    \label{fig1a}
\end{minipage}
\hspace{0.05\columnwidth} 
\begin{minipage}[b]{0.46\columnwidth}
    \centering
    \includegraphics[width=\linewidth]{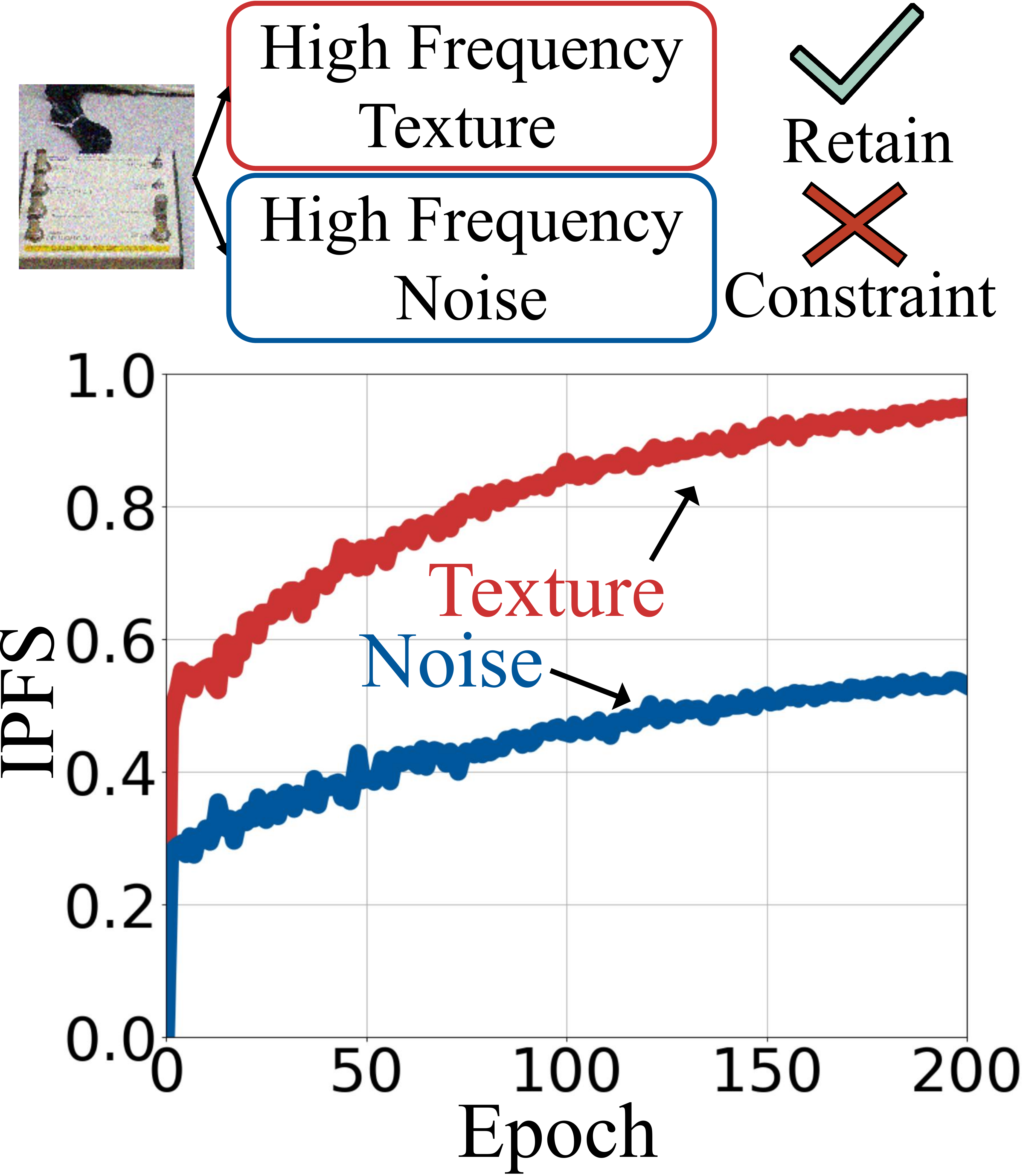} 

    \caption*{(b) Our method} 
    \label{fig1b}
\end{minipage}
\caption{Comparison between existing self-supervised denoising methods and our methods in high frequency. Our method can selectively suppress high-frequency noise while preserving high-frequency texture details.}
\label{fig:com}
\end{figure}
To address the dependency on high-quality noisy/clean image pairs, self-supervised methods have emerged to circumvent the constraints of paired data. DIP \cite{Lempitsky_Vedaldi_Ulyanov_2018} developed methods that learn deep priors from a single noisy image. Masking methods such as \cite{Batson_Royer_2019, Krull_Buchholz_Jug_2018, li2025interpretable,interpretable, chen2025spatiotemporal}, avoid the complexities associated with training using a single image. However, due to the large blind spots in the input, the receptive field for predicting pixels loses a significant amount of valuable context, leading to poorer performance. Noise2Noise (N2N) \cite{Lehtinen_Munkberg_Hasselgren_Laine_Karras_Aittala_Aila_2018}, which theoretically suggests that clean image pairs are not necessary for training, indicates that training solely on noisy images can yield results comparable to those obtained from noisy-clean image pairs. The practical implementation of N2N requires multiple noisy samples of the same scene, which presents challenges in real-world scenarios. LD-RPS~\cite{ld} utilizes the diffusion prior to generate restored image, while ignoring the frequency loss. NBR2NBR and Prompt-SID \cite{Huang_Li_Jia_Lu_Liu_2021, prompt-sid} addresses this challenge by generating sub-image pairs from downsampled versions of the original image. However, these methods that utilize image pairs do not address the spectral bias of the denoising network.
CRnet \cite{yang2024crnet} enhances image quality by unifying restoration. In the unsupervised domain, MCSB \cite{Shi_Mettes_Maji_Snoek_2022} optimizes image restoration with Deep Image Prior (DIP) by controlling the frequency. However, it only works for single image.
 \begin{figure}[t]
 \centering
 \includegraphics[width=0.9\columnwidth]{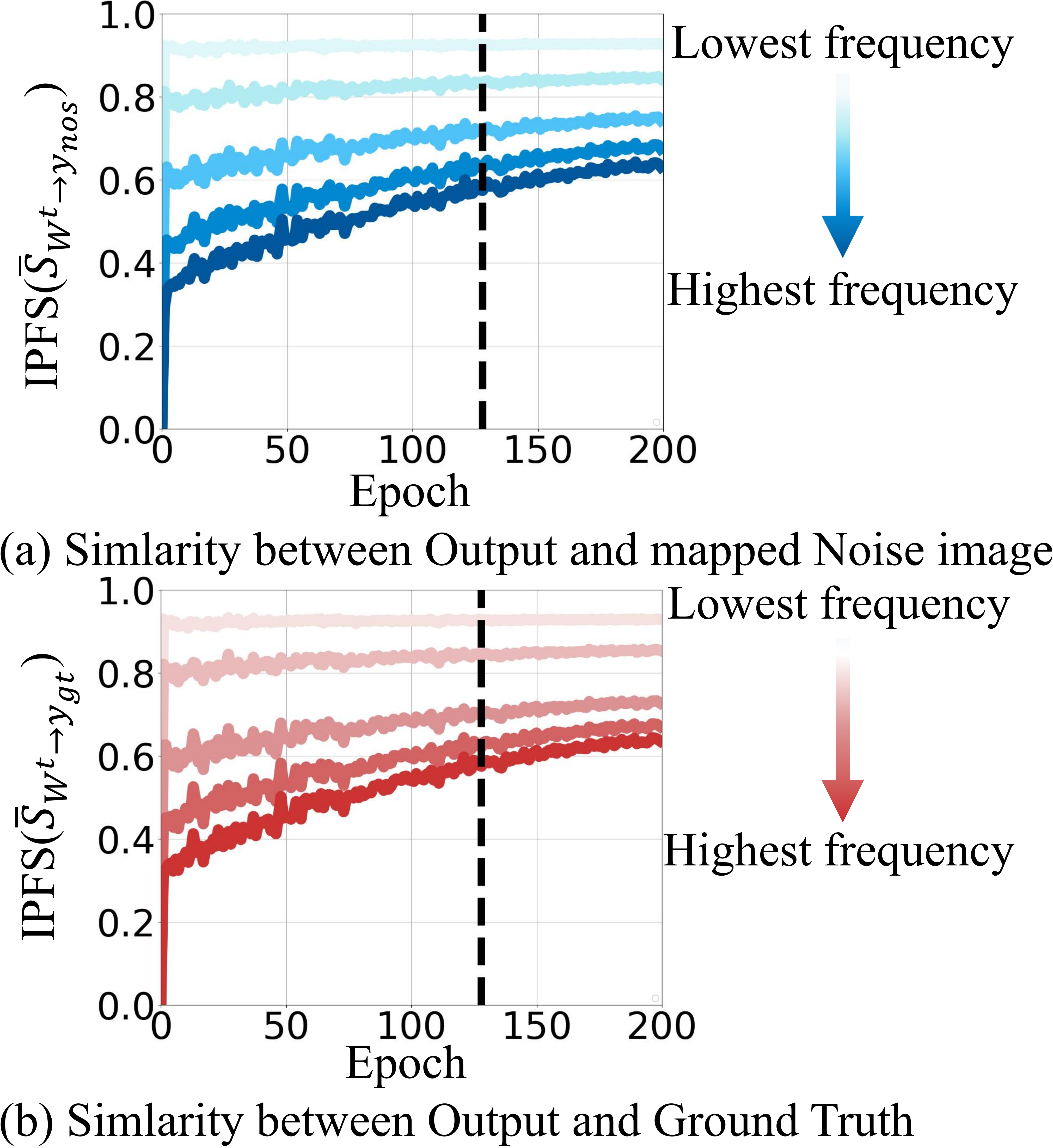} 
 \caption{Spectrum measurement of paired noisy images during network iteration in the existing method. (a) represents the frequency band similarity between the output and the mapped noisy image; (b) represents the frequency band similarity between the output and the ground truth (GT). The peak value of PSNR is indicated by the dashed line.}
 \label{Spectrum measurement}
\end{figure}

\begin{figure*}[!th]
  \centering
  \includegraphics[width=0.9\textwidth]{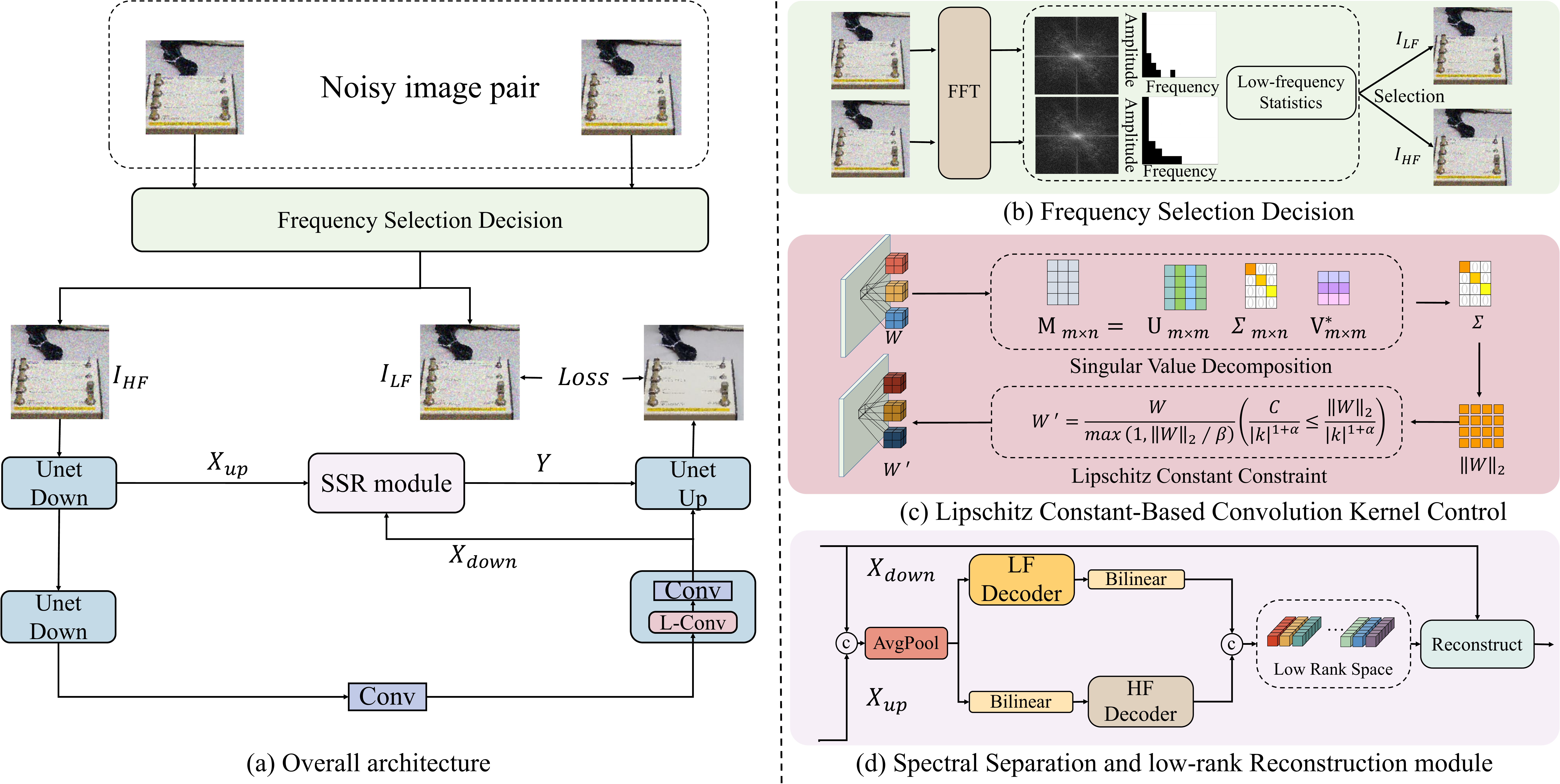}

  \caption{(a) Overall architecture of the SCNet. (b) Perform frequency domain selection on noisy image pairs using FSD. (c) Control convolutional layers to constrain high-frequency noise by using the Lipschitz constant. (d) Separate and reconstruct high-frequency details by using the SSR module.}
  \vspace{0mm}
  \label{fig:oa}
\end{figure*}

As shown in Figure \ref{fig:com}(a), existing methods fail to consider the heterogeneity between high-frequency noise and details when addressing spectral bias, leading the network to treat all high-frequency signals equally, which is detrimental to image content restoration. In contrast, the method presented in Figure \ref{fig:com}(b) effectively addresses this shortcoming. Our approach selectively constrains high-frequency noise in the frequency domain while preserving high-frequency textures, resulting in high-quality self-supervised denoising.

\section{Motivation}
In this work, we demonstrate that the frequency-domain learning capability of the network is crucial for optimizing self-supervised denoising of paired noisy images, which are generated by down-sampling from NBR2NBR. We achieve this by using a metric called Image Pair Frequency-band Similarity (IPFS), which provides spectral measurements between output and mapped noisy images and between output and ground truth images in self-supervised denoising of paired noisy images. the result is shown in Figure \ref{Spectrum measurement}.

The proposed  metric IPFS examines the input-output similarity of multiple frequency bands in the training of NBR2NBR. The calculation of this metric is similar to the Frequency-Band Correspondence (FBC) metric in MCSB \cite{Shi_Mettes_Maji_Snoek_2022}. 
However, our metric measures the frequency band similarity between the network output and the mapped noisy image, as well as between the network output and the ground truth.


\par
As depicted in Figure \ref{Spectrum measurement}, the frequency band similarity between the network output and the mapped noisy image shows that during the network's iterative process, the similarity of low-frequency components reaches its peak initially. In contrast, the similarity of high-frequency components continues to rise throughout the entire iteration process.
During the iterative process, when the network reaches the highest PSNR point, the similarity of high-frequency components between the network output and the ground truth of the noisy images continues to increase while the PSNR does not improve further. At the same time, the similarity of high-frequency components between the network output and the mapped noisy images also continues to increase.

In conclusion, the key to improving the self-supervised denoising effect of paired noisy images is to enhance the network's ability to retain structures of high-frequency signals and constraint the fitting of high-frequency noise.
In this paper, we propose an end-to-end self-supervised denoising framework named SCNet. The main contributions can be summarized as follows:
\begin{itemize}
    \item We suggest choosing images with more high-frequency components as the denoised images pass through the network to enhance the network's ability to extract high-frequency image details and accelerate convergence.   
    \item We propose a self-supervised approach to denoising achieved by controlling the learning rate for high-frequency noisy signals from a new parameter optimization perspective by limiting the Lipschitz constant of convolutional layers.
    \item To retain high-frequency structured textures, we introduce a Spectral Separation and low-rank space Reconstruction module (SSR module). It can be flexibly integrated into the network and separates high-frequency structural information from noise by performing frequency-domain separation and reconstruction. 
    \item Compared to state-of-the-art methods on widely used synthetic and real-image denoising benchmarks, our method delivers highly competitive results.
\end{itemize}
\section{Method}
 In this section, we will provide a detailed description of the workflow for the self-supervised denoising Spectral Control Network (Figure \ref{fig:oa}) we proposed.

\subsection{Overall Architecture}
As is shown in Figure \ref{fig:oa}(a), our SCNet adopts the simple U-Net architecture as the baseline. We first propose a Frequency Selection Decision (FSD), which suggests that noisy images passing through the network should be selectively chosen in the frequency domain. 
Then, we uniformly restrict the network's learning capability for high-frequency signals by controlling the Lipschitz constant of the convolutional kernels, aiming to maximize the denoising effect in the frequency domain.
Finally, to supplement and preserve high-frequency texture information, we use the SSR module to separate textures and noise and retain high-frequency texture details between the upstream and downstream of the network.

The proposed Frequency Selection Decision (Figure \ref{fig:oa}(b)) is a frequency domain selection decision for paired noisy images. We first calculate the high-frequency components of the two noisy images and designate those images with more high-frequency signals to pass through the network for denoising. 
This method helps improve the network's ability to extract high-frequency information. 
\subsection{Lipschitz Control for Convolution Kernels}

From a frequency domain perspective, noise is predominantly present in high frequencies, and the network tends to fit high-frequency noise in the later stages of iteration. This problem can be mitigated by adjusting the spectrum of learnable layers, as shown in Figure \ref{fig:oa}(c), to control the learning rate of these high-frequency signals, we constrain the Fourier coefficients of the learnable layers (i.e., convolutional layers) by limiting the network's Fourier spectrum.
Specifically, We assume that the convolutional layers in our network are Lipschitz continuous. This means that the rate at which the network's output changes in response to its input does not exceed a fixed rate $L$. Formally, for any arbitrary input $x_{1}$ and $x_{2}$, the output $f(x_1)$ and $f(x_2)$ satisfies the following condition:

\begin{equation}
    \|f(x_1)-f(x_2)\|\leq L\|x_1-x_2\|
\end{equation}

where $L$ is the Lipschitz constant. This inequality ensures that the output difference is bounded by the input difference scaled by $L$, indicating that the network's response to changes in the input is controlled and predictable.
The Lipschitz constant $C$ is the smallest value that satisfies the above inequality. It represents the maximum rate at which the function $f$ can change with respect to its input. In other words, $C$ is the smallest bound on the gradient of the function.
\par
The rate of decay of Fourier coefficients is closely related to the smoothness of the function. Smooth functions tend to have Fourier coefficients that decay rapidly, while non-smooth functions have Fourier coefficients that decay more slowly. 
For a Lipschitz continuous function, its Fourier coefficients typically satisfy a certain decay condition:
\begin{equation}
    \|\hat{f}(k)\|\leq\frac C{|k|^{1+\alpha}},
\end{equation}
where $\hat{f}(k)$ represents the Fourier coefficient at frequency $k$, and $\alpha $ is related to the degree of smoothness of the function, the value of $\alpha $ typically ranges from 0 to 1. 
\par
Further, the Lipschitz constant of a convolution layer is bounded by the spectral norm of its parameters, which means $C\leq\|W\|_2$.  We obtain:
\begin{equation}
    \|\hat{f}(k)\|\leq\frac C{|k|^{1+\alpha}}\leq\frac{\|W\|_2}{|k|^{1+\alpha}},
\end{equation}
where $C$ denotes the Lipschitz constant of the convolution layer and $\|W\|_2$ represents the spectral norm of the weight matrix $W$ of the convolution layer.$\|W\|_2$ can be approximated relatively quickly through a few iterations of the power method within the neural network.
\par
The power law $|k|^{1+\alpha}$ indicates that the spectral decay is more pronounced at higher frequencies, implying that learning higher frequencies necessitates a larger spectral norm. Thus, we can control the ability of convolution kernels to learn higher frequencies by constraining its spectral norm $W$ to $\frac{W}{\max(1, \|W\|_2/\beta)}$ with a value $\beta$.
Then, we normalize $W$ by $\|W\|_2/\beta$ if the weight matrix $W$ is higher than $\beta$.

\subsection{Spectral Separation and low-rank Reconstruction}

To retain the high-frequency textures of the original image, we separate the image information into high and low frequency components, enhancing each separately.Figure \ref{fig:oa}(d) shows the details of the SSR module.


We propose a simple and efficient method to separate low-frequency and high-frequency regions and to enhance the frequency domain features independently: First, for a feature  $F$, we use average pooling to downsample it, thereby extracting its low-frequency features $F_{low}^{B\times C\times\frac{H}{2}\times\frac{W}{2}}$.

\begin{table*}[!ht]
  \centering
  \small
  \setlength\tabcolsep{4pt} 
   \resizebox{0.9\textwidth}{!}{
    \begin{tabular}{cllllcllll}
    \toprule
    \multicolumn{1}{l}{Noise Type} & Method & KODAK & BSD300 & SET14 & \multicolumn{1}{l}{Noise Type} & Method & KODAK & BSD300 & SET14 \\
    \midrule
   \multirow{14}{*}{\shortstack[c]{Gaussian\\$\sigma=25$}} 
 & Baseline, N2C & 32.43/0.884 & 31.05/0.879 & 31.40/0.869 &  \multirow{14}{*}{\shortstack[c]{Gaussian\\$\sigma\in[5,50]$}} 
& Baseline, N2C & 32.51/0.875 & 31.07/0.866 & 31.41/0.863 \\
          & Baseline, N2N & 32.41/0.884 & 31.04/0.878 & 31.37/0.868 &       & Baseline, N2N & 32.50/0.875 & 31.07/0.866 & 31.39/0.863 \\
\cmidrule{2-5}\cmidrule{7-10}          & CBM3D & 31.87/0.868 & 30.48/0.861 & 30.88/0.854 &       & CBM3D & 32.02/0.860 & 30.56/0.847 & 30.94/0.849 \\
          & Self2Self & 31.28/0.864 & 29.86/0.849 & 30.08/0.839 &       & Self2Self & 31.37/0.860 & 29.87/0.841 & 29.97/0.849 \\
          & N2V   & 30.32/0.821 & 29.34/0.824 & 28.84/0.802 &       & N2V   & 30.44/0.806 & 29.31/0.801 & 29.01/0.792 \\
          & Laine19-mu & 30.62/0.840 & 28.62/0.803 & 29.93/0.830 &       & Laine19-mu & 30.52/0.833 & 28.43/0.794 & 29.71/0.822 \\
          & Laine19-pme & \underline{32.40/0.883} & 30.99/\underline{0.877} & \underline{31.36/0.865} &       & Laine19-pme & \underline{32.40}/0.870 & 30.95/0.861 & \underline{31.21}/0.855 \\
          & DBSN  & 31.64/0.856 & 29.80/0.839 & 30.63/0.846 &       & DBSN  & 30.38/0.826 & 28.34/0.788 & 29.49/0.814 \\
          & R2R   & 32.25/0.880 & 30.91/0.872 & 31.32/0.865 &       & R2R   & 31.50/0.850 & 30.56/0.855 & 30.84/0.850 \\
          & NBR2NBR & 32.08/0.879 & 30.79/0.873 & 31.09/0.864 &       & NBR2NB & 32.10/0.870 & 30.73/0.861 & 31.05/\underline{0.858} \\
          & Blind2Ublind & 32.27/0.880 & 30.87/0.872 & 31.27/0.864 &       & Blind2Ublind & 32.34/\underline{0.872} & 30.86/{0.861} & 31.14/0.857 \\
          & swinIA & 30.12/0.819 & 28.40/0.789 & 29.54/0.814 &       & swinIA & 30.30/0.820 & 28.40/0.785 & 29.49/0.809 \\
          & DCD-net & 32.27/0.881 & \underline{31.01}/0.876 & 31.29/0.862 &       & DCD-net & 32.35/\underline{0.872} & \underline{31.09/0.866} & 31.09/0.855 \\
          & SCNet (Ours)  & \textbf{32.50}/\textbf{0.886} & \textbf{31.19}/\textbf{0.882} & \textbf{31.55/0.871} &       & SCNet (Ours)  & \textbf{32.60}/\textbf{0.878} & \textbf{31.23}/\textbf{0.871} & \textbf{31.54/0.867} \\
    \midrule
     \multirow{14}{*}{\shortstack[c]{Poisson\\$\lambda=30$}} 
 & Baseline, N2C & 31.78/0.876 & 30.36/0.868 & 30.57/0.858 &  \multirow{14}{*}{\shortstack[c]{Poisson\\$\lambda\in[5,50]$}}  & Baseline, N2C & 31.19/0.861 & 29.79/0.848 & 30.02/0.842 \\
          & Baseline, N2N & 31.77/0.876 & 30.35/0.868 & 30.56/0.857 &       & Baseline, N2N & 31.18/0.861 & 29.78/0.848 & 30.02/0.842 \\
\cmidrule{2-5}\cmidrule{7-10}          & Anscombe & 30.53/0.856 & 29.18/0.842 & 29.44/0.837 &       & Anscombe & 29.40/0.836 & 28.22/0.815 & 28.51/0.817 \\
          & Self2Self & 30.31/0.857 & 28.93/0.840 & 28.84/0.839 &       & Self2Self & 29.06/0.834 & 28.15/0.817 & 28.83/0.841 \\
          & N2V   & 28.90/0.788 & 28.46/0.798 & 27.73/0.774 &       & N2V   & 28.78/0.758 & 27.92/0.766 & 27.43/0.745 \\
          & Laine19-mu & 30.19/0.833 & 28.25/0.794 & 29.35/0.820 &       & Laine19-mu & 29.76/0.820 & 27.89/0.778 & 28.94/0.808 \\
          & Laine19-pme & \underline{31.67/0.874} & \underline{30.25/0.866} & \underline{30.47/0.855} &       & Laine19-pme & 30.88/0.850 & 29.57/0.841 & 28.65/0.785 \\
          & DBSN  & 30.07/0.827 & 28.19/0.790 & 29.16/0.814 &       & DBSN  & 29.60/0.811 & 27.81/0.771 & 28.72/0.800 \\
          & R2R   & 30.50/0.801 & 29.47/0.811 & 29.53/0.801 &       & R2R   & 29.14/0.732 & 28.68/0.771 & 28.77/0.765 \\
          & NBR2NBR & 31.44/0.870 & 30.10/0.863 & 30.29/0.853 &       & NBR2NBR & 30.86/0.855 & 29.54/0.843 & 29.79/0.838 \\
          & Blind2Ublind & 31.64/0.871 & \underline{30.25}/0.862 & 30.46/0.852 &       & Blind2Ublind & \underline{31.07/0.857} & 29.92/0.852 & \underline{30.10/0.844} \\
          & swinIA & 29.51/0.805 & 27.92/0.775 & 28.74/0.799 &       & swinIA & 29.06/0.788 & 27.74/0.764 & 28.27/0.780 \\
          & DCD-net & 31.60/0.870 & 30.22/0.865 & 30.41/\underline{0.855} &       & DCD-net & 31.00/\underline{0.857} & \underline{29.99/0.855} & 29.99/0.843 \\
          & SCNet (Ours)  & \textbf{31.88}/\textbf{0.879} & \textbf{30.55}/\textbf{0.872} & \textbf{30.77/0.861} &       & SCNet (Ours)  & \textbf{31.35}/\textbf{0.865} & \textbf{30.19}/\textbf{0.860} & \textbf{30.48/0.852} \\
    \bottomrule
    \end{tabular}%
    }
    \caption{Quantitative denoising results on synthetic datasets in sRGB space. The highest PSNR(dB)/SSIM among unsupervised denoising methods is highlighted in \textbf{bold}, while the second is \underline{underlined}.}
  \label{tab:sRGB}%
\end{table*}%
Secondly, we use bilinear interpolation to upsample $F_{low}$ to the original dimensions $B\times C\times H\times W$. Then, by subtracting these upsampled features from the original features, we obtain the high-frequency information of the original feature, denoted as $F_{high}$. 
where Upsample refers to the up-sampling operation using bilinear interpolation.
Finally, we upsample the decoded low-frequency features and concatenate them with the decoded high-frequency information to generate the basis vectors for the image's low-rank space.
\begin{figure*}[!h]
  \centering
  \includegraphics[width=1\textwidth]{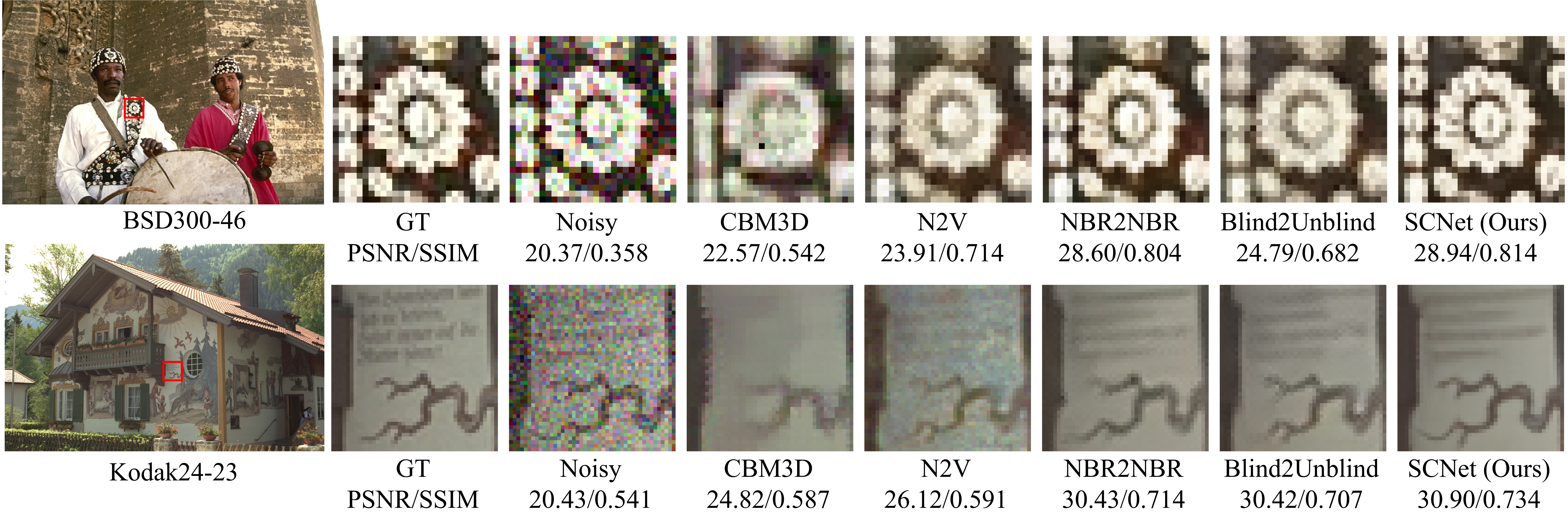}

  \caption{Visual comparison of denoising sRGB images in the setting of $\sigma=25$.}
  \label{fig:srgb}
\end{figure*}
\subsubsection{Low-Rank Space Reconstruction.}
For the generation of the low-rank space and the reconstruction of high-frequency signals there are two steps.
Take two feature maps of the same dimensions from both the upstream and downstream of the network, derived from the same image, denoted as $X_{up}$ and $X_{down}$, and concatenate them to generate the basis extraction vector $X_{E}\in\mathbb{R}^{B\times C\times H\times W}$ .
The  low-rank basis vector Generator for reconstruction: $G_{\theta} : \mathbb{R}^{B\times C\times H\times W} \to \mathbb{R}^{B\times k\times N}$ be a function parameterized by $\theta$ can denoted as:
\begin{equation}
   v_{1},v_{2},\cdots,v_{k}=G_\theta(X_{E}),
\end{equation}
where $X_{E}$  is a feature map derived from the high-frequency enhancement module, and $v_{1},v_{2},\cdots,v_{k}$ be the $k$ low-rank basis vectors. We define a function $G_\theta$ implemented by a convolutional block. We input $X_{E}$ into a simple convolutional block with $k$ output channels to obtain $k$ low-rank basis vectors for the reconstruction of $X_{down}$.
\par
Since the $k$ basis vectors of the low-rank space generated automatically by the neural network cannot be guaranteed to be orthogonal to each other, an orthogonalization operation is necessary: let matrix $V=[v_{1},v_{2},\cdots,v_{k}]^\intercal\in\mathbb{R}^{N\times k}$, the column vectors of $V$  represent the basis vectors of the $k$-dimensional low-rank space $\mathcal{V}\subset\mathbb{R}^{N}$.\par
Let $P:\mathbb{R}^N\to\mathcal{V}$, $P$ can be calculated from $V$:
\begin{equation}
    P=V(V^\intercal  V)^{-1}V^\intercal ,
\end{equation}
where the $P$ is the orthogonal projection matrix of the input image in the low-rank space. The normalization term $(V^\intercal V)^{-1}$ is required for orthogonalization.\par
Then, we use the orthogonal projection matrix $P$ to reconstruct $X_{down}$ in the high-frequency enhanced low-rank space, obtaining a denoised feature map with enhanced high-frequency components, given by:
\begin{equation}
    Y=PX_{down},
\end{equation}
where the $Y$ is the reconstructed by SSR module from downstream feature map. The reconstruction operations are fully differentiable, thus enabling learning within the network.

\section{Experiments}
We evaluate our framework on various denoising tasks with both synthetic and real-world noise.

\subsection{Synthetic Noise}
Following the setup of previous methods, we selected the ILSVRC2012 \cite{deng2009imagenet} validation dataset with image sizes ranging from 256x256 to 512x512 pixels as our validation set. We tested our results on Kodak, BSD300, and Set14, with 10, 3, and 20 iterations respectively, in line with previous experiments. Thus, all methods were evaluated using 240, 300, and 280 individual synthetic noisy images. Specifically, we consider four types of noise in sRGB space: (1) Gaussian noise with $\sigma=25$, (2) Gaussian noise with $\sigma \in [5,50]$, (3) Poisson noise with $\lambda=30$, (4) Poisson noise with $\lambda \in [5,50]$. 
As shown in Table\ref{tab:sRGB}, traditional methods such as BM3D perform worse than deep learning methods, with a significant gap. This indicates that neural networks capture image priors better and fit the distribution of natural images more effectively than traditional handcrafted models. Additionally, other self-supervised denoising methods also demonstrate good denoising results. Despite their effectiveness, they fall significantly short of supervised denoising methods. Our method outperforms model-based or other self-supervised methods both qualitatively and quantitatively. As shown in Figure \ref{fig:srgb}, SCNet demonstrates promising denoising performance across various types of noise. The denoised images are clean and sharp. More importantly, it does not rely on paired data or known noise statistics, highlighting its potential value in applications.
\begin{table}[ht]
\scriptsize
  \centering
  \setlength{\abovecaptionskip}{0.1cm} 
  \setlength\tabcolsep{7pt}
  \footnotesize
  \resizebox{1\columnwidth}{!}{
  \begin{tabular}[b]{@{}cccc@{}}
    \toprule
      \multirow{2}{*}{Methods} 
      & Confocal & Confocal & Two-Photon\\
      &Fish &Mice &Mice\\
    \midrule
      Baseline, N2C  & 32.79/0.905 & 38.40/0.966 & 34.02/0.925 \\
      Baseline, N2N  & 32.75/0.903 & 38.37/0.965 & 33.80/0.923 \\
    \midrule
      BM3D  & 32.16/0.886 & 37.93/0.963 & 33.83/\textbf{0.924} \\
      N2V & 32.08/0.886 & 37.49/0.960 & 33.38/0.916 \\
      Laine19-mu (G)  & 31.62/0.849 & 37.54/0.959 & 32.91/0.903 \\
      Laine19-pme (G)  & 23.30/0.527 & 31.64/0.881 & 25.87/0.418 \\
      Laine19-mu (P) & 31.59/0.854 & 37.30/0.956 & 33.09/0.907 \\
      Laine19-pme (P)  & 25.16/0.597 & 37.82/0.959 & 31.80/0.820 \\
      NBR2NBR& 32.11/0.890 & 37.07/0.960 & 33.40/0.921 \\
      Blind2Unblind  & 32.74/0.897 &38.44/\textbf{0.964} & 34.03/0.916 \\
      SCNet (Ours) & \textbf{33.10/0.910} & \textbf{38.84/0.964} & \textbf{34.16/0.924} \\
    \bottomrule
  \end{tabular}
  }
  \caption{Comparison on Confocal Fish, Confocal Mice and Two-Photon Mice. G is for Gaussian and P is for Poisson.}
  \label{tab:fmd}
\end{table}
\begin{figure}[!h]
  \centering
  \includegraphics[width=1\columnwidth]{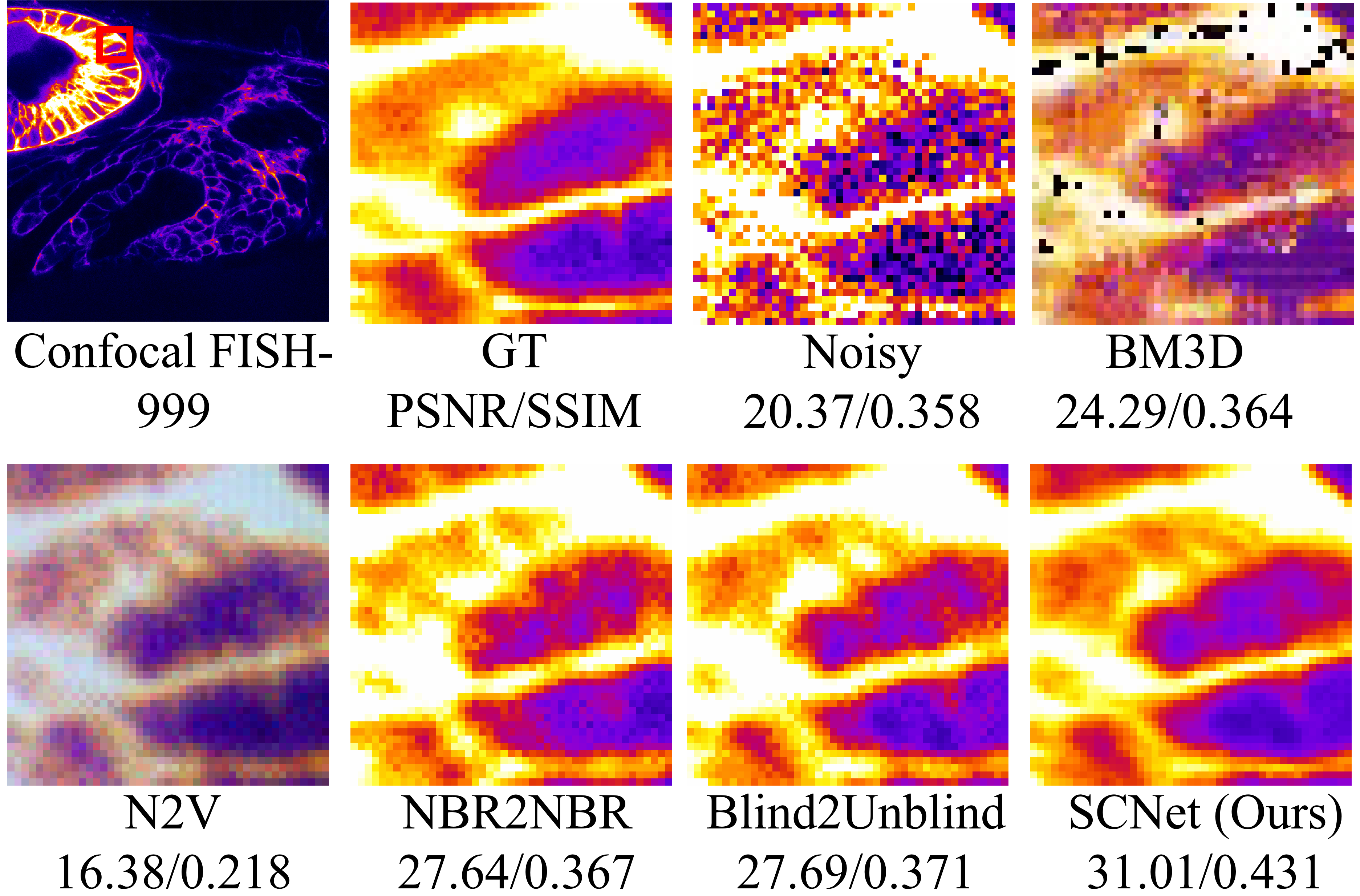}
  \caption{Visual comparison of denoising FM images.}
  \label{fig:Vfm}
\end{figure}

\subsection{Real-world Noise} We further evaluated the performance of our pipeline on the real fluorescence microscopy denoising (FMD) dataset\cite{zhang2019poisson}. The noisy microscopy images in this dataset were acquired using commercial confocal, two-photon, and wide-field microscopes. 
Compared to other methods, our approach demonstrates competitive performance. In particular, our method significantly outperforms self-supervised methods and shows slight improvements over supervised methods (N2C and N2N) in all sub-datasets. We present the quantitative results in Table \ref{tab:fmd}. The visual comparisons can be found in Figure \ref{fig:Vfm}.

  \begin{table}[ht]\normalsize
    \centering
    \setlength{\tabcolsep}{4pt}
    \resizebox{1\columnwidth}{!}{
    \begin{tabular}[b]{@{}cccccccc@{}}
    \toprule
        method & baseline & FSD & Lipschitz & SSR & PSNR & SSIM \\ \midrule
        I & $\checkmark$ & ~ & ~ & ~ & 31.09 & 0.864 \\ 
        II & $\checkmark$ & $\checkmark$ & ~ & ~ & 31.13 & 0.865 \\ 
        III & $\checkmark$ & ~ & $\checkmark$ & ~ & 31.22 & 0.865 \\ 
        IV & $\checkmark$ & ~ & ~ & $\checkmark$ & 31.41 & 0.867 \\ 
        V & $\checkmark$ & ~ & $\checkmark$ & $\checkmark$ & 31.50 & 0.869 \\ 
        VI & $\checkmark$ & $\checkmark$ & $\checkmark$ & $\checkmark$ & \textbf{31.55} & \textbf{0.871} \\ \bottomrule
    \end{tabular}
    }

     \caption{Analysis of different component combinations.}
  \label{tab:ab}

\end{table}

\subsection{Ablation Study} This section conducts an ablation study of the Frequency Selection Decision, Lipschitz method, and SSR module. Note that PSNR(dB)/SSIM is evaluated on the Set14 dataset.
\subsubsection{Component Analysis.} Table \ref{tab:ab} lists the performance of different component. II outperforms I by 0.04 in PSNR and 0.001 in SSIM. From this, it is evident that FSD improves denoising performance. And FSD accelerates convergence by 23 epochs compared to the baseline during training in Synthetic Noise. III outperforms I by 0.13 in PSNR and 0.002 in SSIM, this indicates that Lipschitz method are highly effective. IV outperforms I by 0.32 in PSNR and 0.003 in SSIM, this indicates that SSR preserves and supplements high-frequency texture details. When used after the Lipschitz method, it further enhances the texture restoration, resulting in a 0.46 dB improvement over the baseline.
\subsubsection{High-frequency Denosing of Lipschitz.} 
\begin{figure}[!t]
  \centering
  \includegraphics[width=0.9\columnwidth]{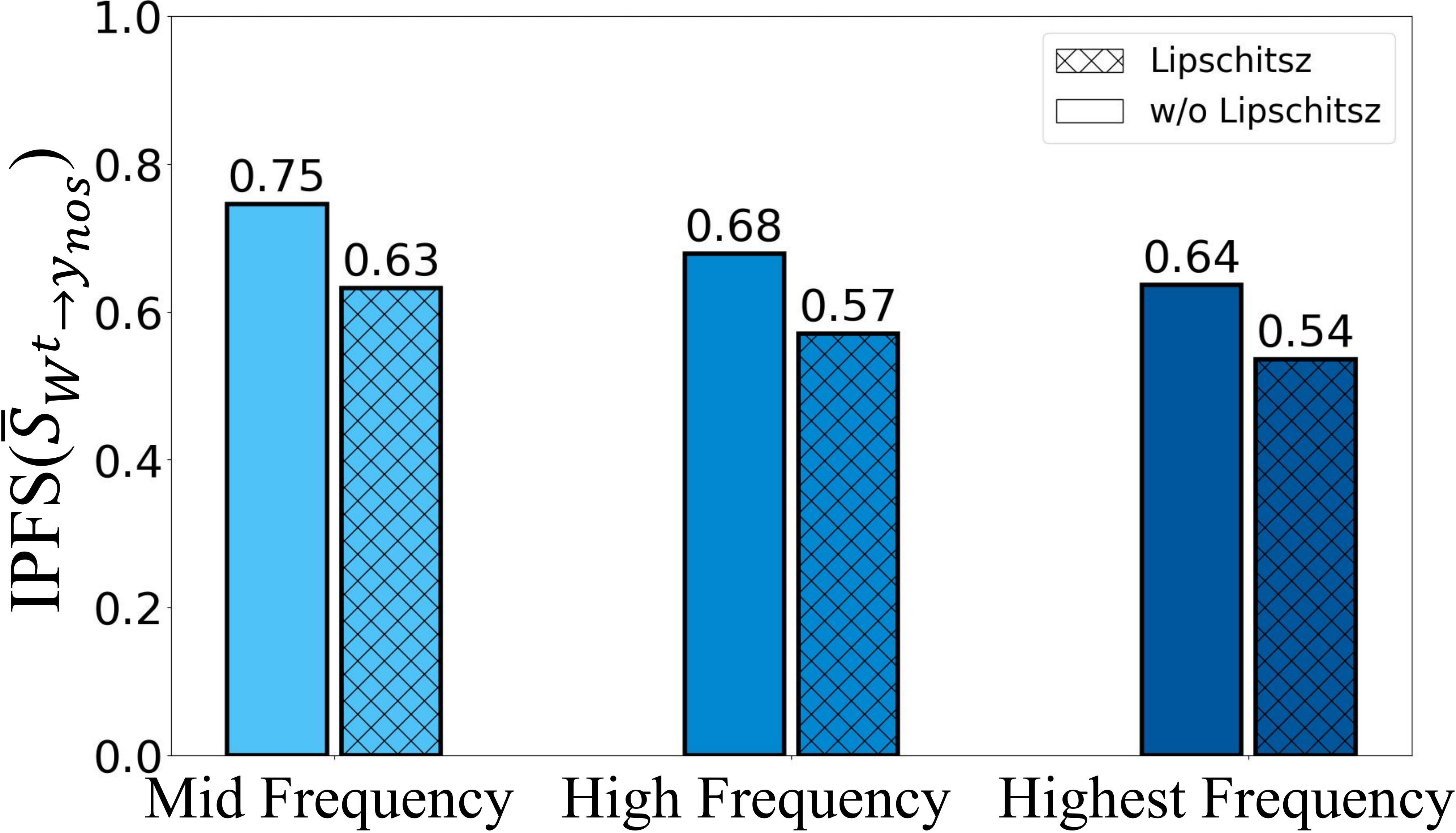}
  \caption{The denoising performance of the Lipschitz method on top 3 frequency bands ($\bar{S}_{W^{(t)} \rightarrow y_{nos}}$, similarity between the output and the noisy image).}
  \label{fig:lm}
\end{figure}

Figure \ref{fig:lm} shows the noise suppression capability of the Lipschitz method across the mid, high, and highest frequency bands. It is evident that there is a decrease in similarity with the noisy image by 0.12, 0.11, and 0.10 in these frequency bands, respectively, indicating that the Lipschitz method effectively suppresses the mapping of high-frequency noise.

\subsubsection{High-frequency Texture Reconstruction of SSR.} 
\begin{figure}[t!h]
  \centering
  \includegraphics[width=1\columnwidth]{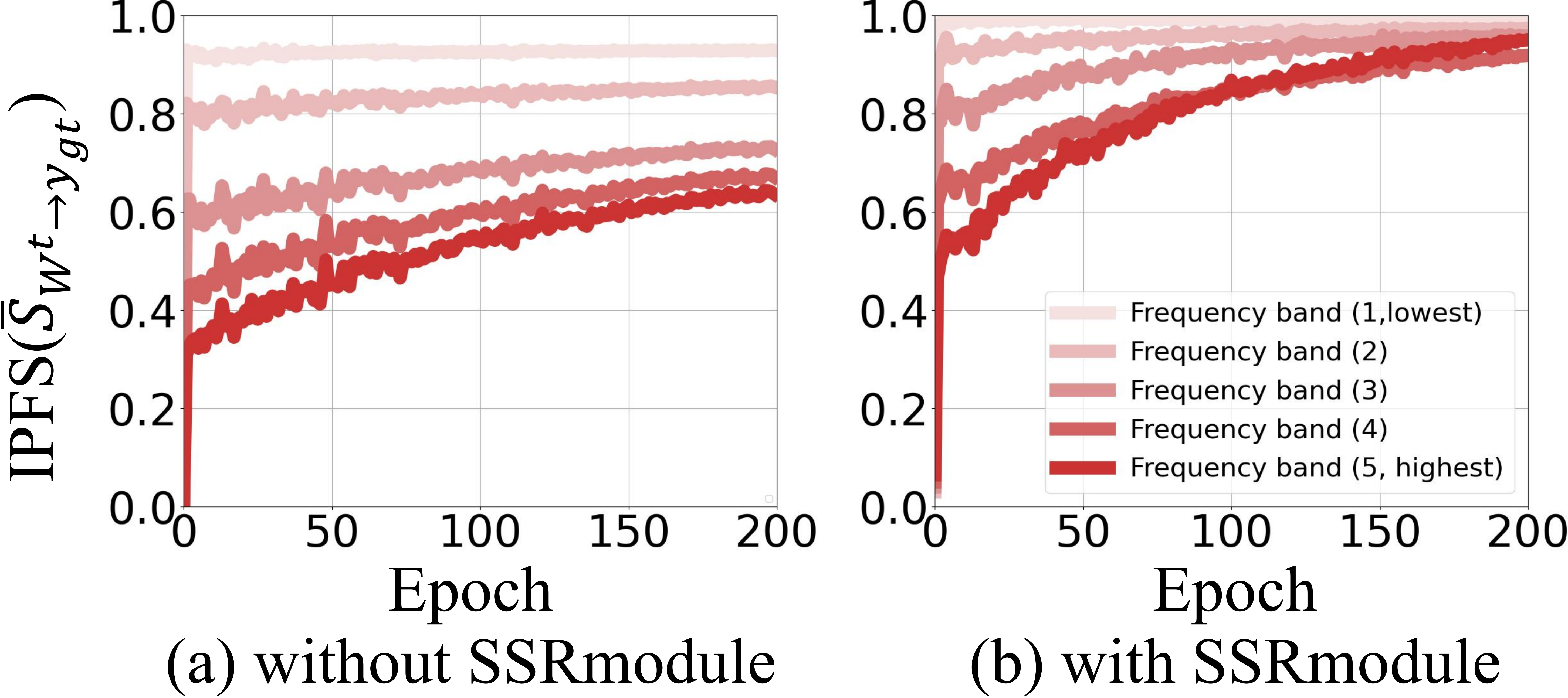}
  \caption{The restoration performance of the SSR module across all frequency bands ($\bar{S}_{W^{(t)} \rightarrow y_{gt}}$, similarity between the output and the ground truth).}
  \label{fig:ssrab}
\end{figure}

Figure \ref{fig:ssrab} illustrates the SSR module's ability to reconstruct frequency domain information during training iterations. It is evident that the SSR module excels at restoring high-frequency band information, achieving a similarity close to 1 with the ground truth across all frequency bands. This represents a significant improvement compared to the baseline.

\section{Conclusion}
We have demonstrated that the challenge with noisy image pairs lies in separating high-frequency texture details from noise. Based on this conclusion, we propose a new framework that separates high-frequency textures from noise in the frequency domain, addressing the spectral bias in self-supervised denoising. We show the effectiveness and broad applicability of the proposed SCNet across various denoising tasks involving different noise attributes and data types. We believe this research highlights the advantages of exploring frequency domain  processing for self-supervised denoising, offering a new perspective of spectral bias in this field.

\bibliographystyle{IEEEbib}
\bibliography{icme2025references}

\end{document}